\documentclass{ieeeaccess}
\usepackage{cite}
\usepackage{amsmath,amssymb,amsfonts}
\usepackage{algorithmic}
\usepackage{graphicx}
\usepackage{textcomp}
\def\BibTeX{{\rm B\kern-.05em{\sc i\kern-.025em b}\kern-.08em
    T\kern-.1667em\lower.7ex\hbox{E}\kern-.125emX}}
\begin{document}
\history{}
\doi{}

\title{A Waste Copper Granules Rating System Based on Machine Vision}
\author{\uppercase{KAIKAI ZHAO}\authorrefmark{},
\uppercase{YAJIE CUI\authorrefmark{}, ZHAOXIANG LIU\authorrefmark{}, and SHIGUO LIAN\authorrefmark{}
}}
\address[]{AI Innovation and Appliaction Center, China Unicom Digital Technology Co., Ltd, Beijing 100013, China}

\markboth
{K. Zhao \headeretal: A Waste Copper Granules Rating System Based on Machine Vision}
{K. Zhao \headeretal: A Waste Copper Granules Rating System Based on Machine Vision}

\corresp{Corresponding author: ZHAOXIANG LIU (e-mail: liuzx178@chinaunicom.cn), and SHIGUO LIAN (e-mail: liansg@chinaunicom.cn).}

\begin{abstract}
In the field of waste copper granules recycling, engineers should be able to identify all different sorts of impurities in waste copper granules and estimate their mass proportion relying on experience before rating. This manual rating method is costly, lacking in objectivity and comprehensiveness. To tackle this problem, we propose a  waste copper granules rating system based on machine vision and deep learning. We firstly formulate the rating task as a 2D image recognition and purity regression task. Then we design a two-stage convolutional rating network to compute the mass purity and rating level of waste copper granules. Our rating network includes a segmentation network and a purity regression network, which respectively calculate the semantic segmentation heatmaps and purity results of the waste copper granules. After training the rating network on the augmented datasets, experiments on real waste copper granules demonstrate the effectiveness and superiority of the proposed network. Specifically, our system is superior to the manual method in terms of accuracy, effectiveness, robustness, and objectivity. 
\end{abstract}

\begin{keywords}
Machine vision, Purity regression network, Rating network, Semantic segmentation, Waste copper granules rating.
\end{keywords}

\titlepgskip=-15pt

\maketitle

\section{Introduction}
\label{sec:introduction}
\PARstart{C}{opper} is a kind of valuable earth resource. 
Unlike coal, oil and other resources, copper is 100$\%$ recyclable. 
Regardless of how many times it is recycled and refined, copper can always maintain its inherent properties. 
The quality of recycled copper is no different from that of newborn copper, 
which makes it a top recyclable material and one of the metals with the highest recycling rate \cite{b1}. 
Meanwhile, copper recycling will become more and more noteworthy over time due to the lack of copper resources and a rise in demand \cite{b2}.

\Figure[t!](topskip=0pt, botskip=0pt, midskip=0pt){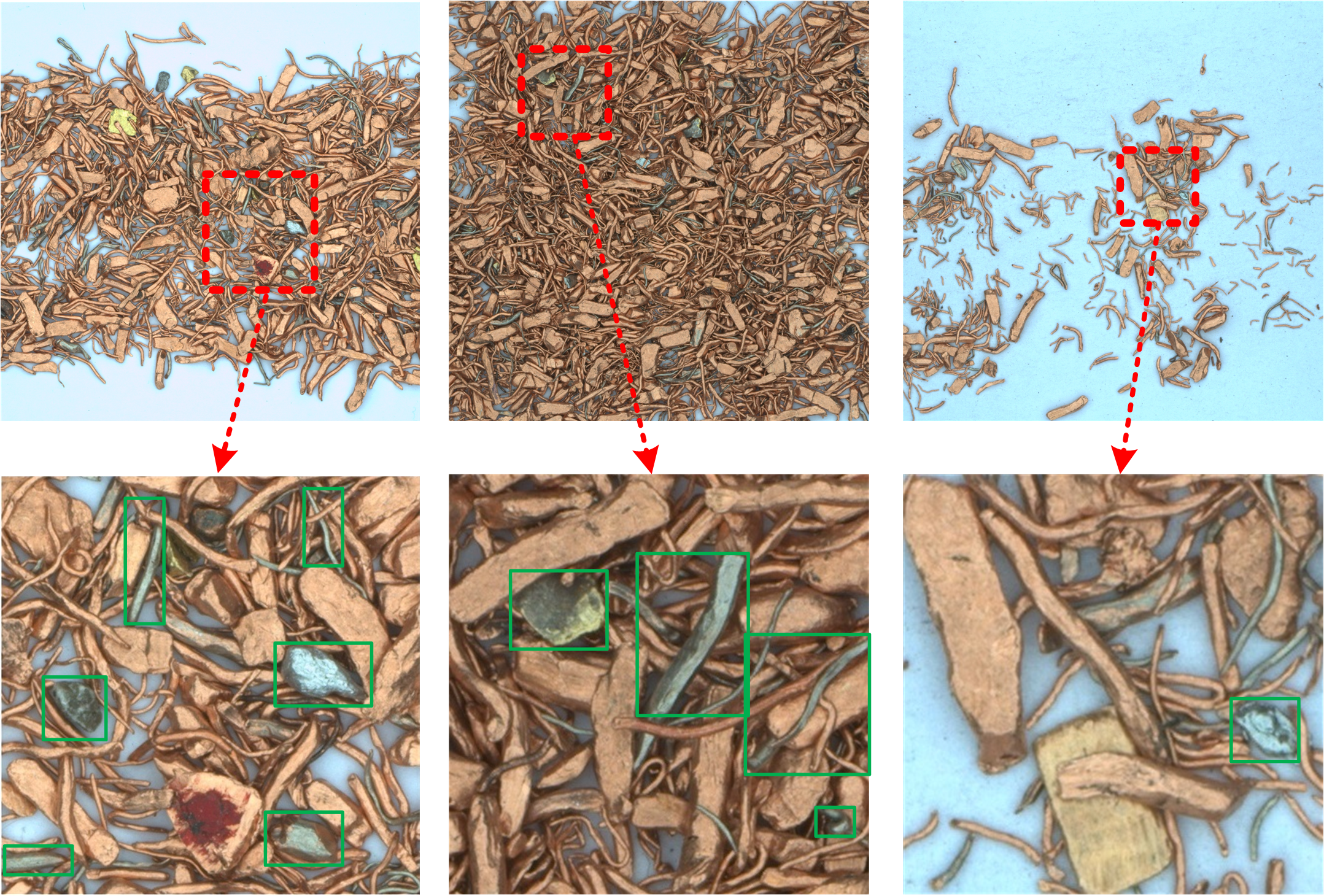}
{Three images of waste copper granules(first row) and their local enlarged view(second row). 
Impurities are marked with green bounding boxes.\label{fig1}}

Waste copper granules is one of the important forms of waste copper. Fig. 1 shows some waste copper granules images and their local enlarged views. Waste copper granules contains a variety of impurities in different proportions besides copper. It’s obvious that the higher the impurity content is, the lower the copper content is and the lower the unit price of recycling is. Therefore, it is necessary to analyze and rate the purity of copper before recycling.

Rating waste copper granules is the process of calculating the mass proportion of copper in waste copper granules and assigning the quality level. We describe the mass proportion of copper as “mass purity” and the mass purity is positively correlated to the price of recycling waste copper granules. Setting varying recycling prices for various levels after manual rating has been the most widely utilised pricing method up to this point. In the manual rating process, professionals use simple equipments to stir and analyse some waste copper granules samples, determine rating level of waste copper granules by visual observation and experience, and then price all waste copper granules to be recycled \cite{b3}. 

To rate waste copper granules accurately and effectively, we propose a waste copper granules rating system based on machine vision and deep learning. Our rating system formulates the waste copper granules rating task into a 2D image recognition and purity regression task, and calculates the purity and rating result of waste copper granules using a convolutional rating network, which consists of a segmentation subnetwork and a purity regression subnetwork. Compared with manual method, our system calculates copper mass purity and outputs rating level in a more efficient and objective way, meanwhile, the proposed system can improve the rating accuracy in an effective way.

The rest of the paper is organized as follows. Section 2 summarizes related works. Section 3 is the main part of the paper, in which we describe the proposed rating process and elaborate the proposed segmentation and purity regression network based on deep learning. To verify the viability of the suggested approach, in Section 4 we conduct extensive experiments on real waste copper granules datasets. Lastly, we give a conclusion and discuss our contributions.

\section{RELATED WORK}
In this section, we introduce related works on the existing manual waste copper granules rating method, industrial inspection and semantic segmentation.

\Figure[t!](topskip=0pt, botskip=0pt, midskip=0pt)[width=1\textwidth]{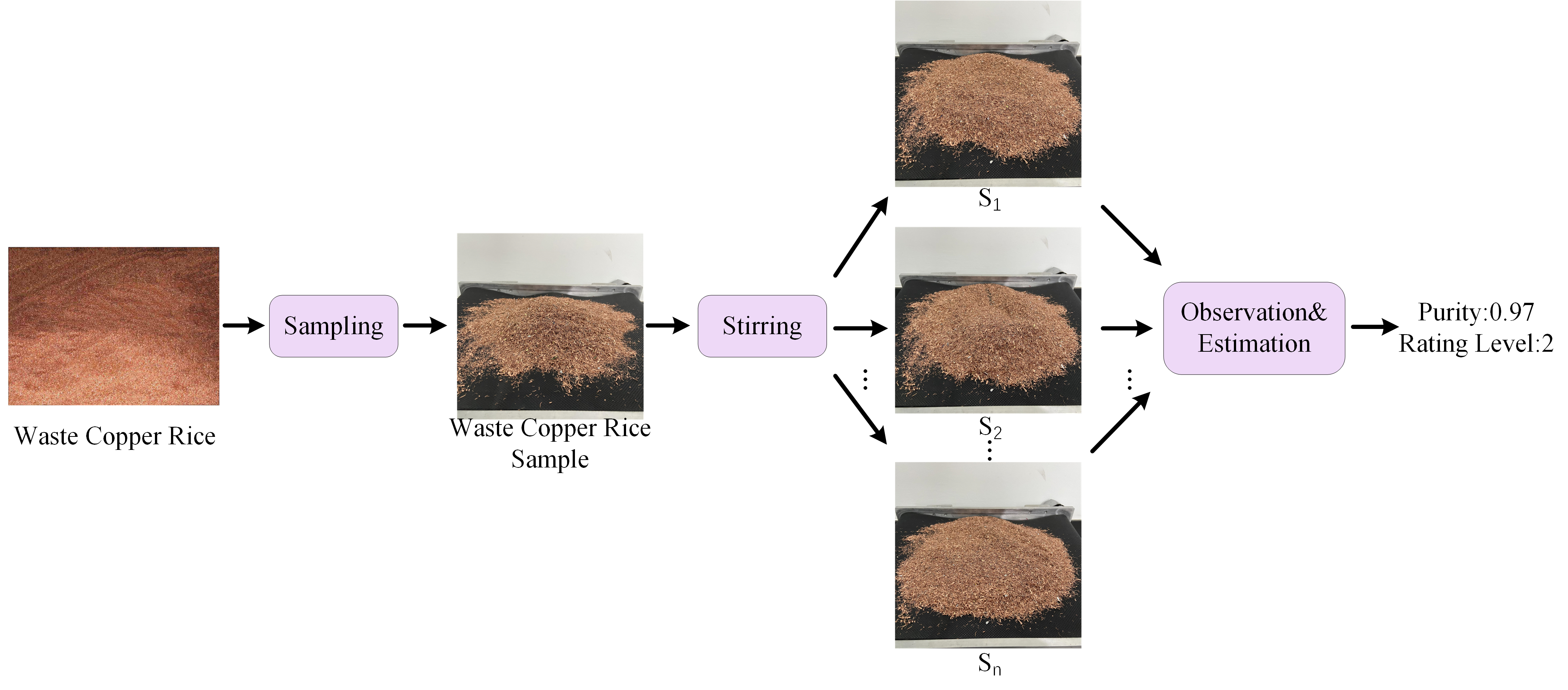}
{Manual rating process. 
After sampling and stirring, engineers observe and estimate the rating result according their experience.\label{fig2}}

\subsection{Manual rating method}
In the field of waste copper granules rating, the existing method mainly relies on manual experience to recognize all kinds of impurities. Different impurities have different densities, so the mass purity of copper will vary depending on the type and amount of impurities present in waste copper granules. The manual rating process is described in Fig. 2. Firstly, it’s necessary to sample the waste copper granules to be recycled because of the large amount of waste copper granules and limited labor. Secondly, engineers have to stir the waste copper granules sample over and over again to observe the sample thoroughly. Finally, engineers estimate the mass purity of copper and the rating level according to their knowledge and expertise.

The existing manual rating method has some obvious shortcomings. Firstly, it can be demanding and laborious to train a professional engineer. And the error induced by manual experience is inevitable. While employing the manual rating approach, the sampling scale is constrained, which further reduces accuracy. Last but not least, the manual technique certainly lacks neutrality because the rating result directly impacts recycling pricing. 

\subsection{Industrial inspection}
In the field of industrial quality inspection, researches pertinent to our task mainly cover three aspects \cite{b4}. Some studies focus on recognizing the types of waste products and do not care about the content of impurities. For example, the method in \cite{b5} detects and classifies waste iron using YOLO, the method in \cite{b6} recognizes wheat using CNN classifier, and the method in \cite{b7} locates defect position in crystalline silicon. Some studies use computer vision methods to analyze the purity of single object with regular shape and distinctive feature. For example, methods in \cite{b8,b9} identify the purity of single corn seed or cotton seed. Other researchers investigate ways to improve chemical technologies for waste product purification. For example, methods in \cite{b10,b11} restore copper using chemical methods. In our waste copper granules rating system, we are committed to calculating the purity of copper, analyzing the quality level of waste copper granules before purification. Due to the variety of impurity types in waste copper granules and the shape of impurities and copper is uncertain and irregular, the current visual methods in industrial inspection cannot solve our problem very well.

\subsection{Semantic segmentation}
Deep learning has achieved significant progress in many visual tasks\cite{b12,b13,b14,b15,b16,b17,b18,b19,b20, b21,b22,b23,b24,b25,b26,b27,b28}, including image classification, object detection, face recognition, semantic segmentation and so on. Considering that the shape of copper granules is irregular and the type of impurity is uncertain, we utilize the deep learning based segmentation model to get the accurate contour of copper in our rating system. Typical semantic segmentation models extract deep semantic features by down sampling and shallow spatial features by maintaining high resolution information, classify each pixel of input image, and output classification heatmap with the same size as input image. Typical semantic segmentation models based on deep learning include UNet\cite{b29}, DeepLabV3+\cite{b30}, NonLocalNet\cite{b31}, CGNet\cite{b32} and Segmenter\cite{b33}, etc. Through extensive data training, a fully convolutional sematic segmentation model can learn, extract, and integrate rich multilevel features to accurately represent the contour of various objects.

\section{Proposed Approach}
\subsection{process of our rating system}
Imitating the manual rating approach, a naive machine vision based method identifies all varieties of impurities in waste copper granules images, evaluates their volume percentage and then computes the mass purity of copper using \eqref{eq1}.
\begin{equation}
\begin{aligned}
{m_0} &= m - \sum\nolimits_{i = 1}^C {\left( {{\rho _i} \times p_i^v \times V} \right)} \\
P &= {m_0}/m
\end{aligned}
\label{eq1}\end{equation}
Here $m_0$ is the mass of copper and $P$ is the mass purity, $m$ is the total mass of waste copper granules, $V$ is the total volume of waste copper granules, and the subscript $i$ which ranges from 1 to $C$ corresponds to C kinds of impurities in waste copper granules. ${\rho _i}$ is the impurity’s density, $p_i^v$ represents the volume proportion of the corresponding impurity. With $m$ and $V$ being determined constant, $p_i^v$ is the variable to be analyzed in the rating task. 

However, in the actual setting, this approach has the following drawbacks. Firstly, the category of impurities is uncertain and it is impossible to enumerate all impurity categories in waste copper granules. Secondly, it is difficult to accurately estimate the density of various impurities. Last but not least, it is too costly to manually label all kinds of impurities in waste copper granules. 

To solve problems of the naive method, we first define all types of impurities as one category, then employ a semantic segmentation model to distinguish between copper and non-copper in each waste copper granules image, and finally design a three-branch purity regression network to calculate the mass purity and rating level accurately, effectively and stably. Our proposed rating process is shown in Fig. 3. 
\Figure[t!](topskip=0pt, botskip=0pt, midskip=0pt)[width=1\textwidth]{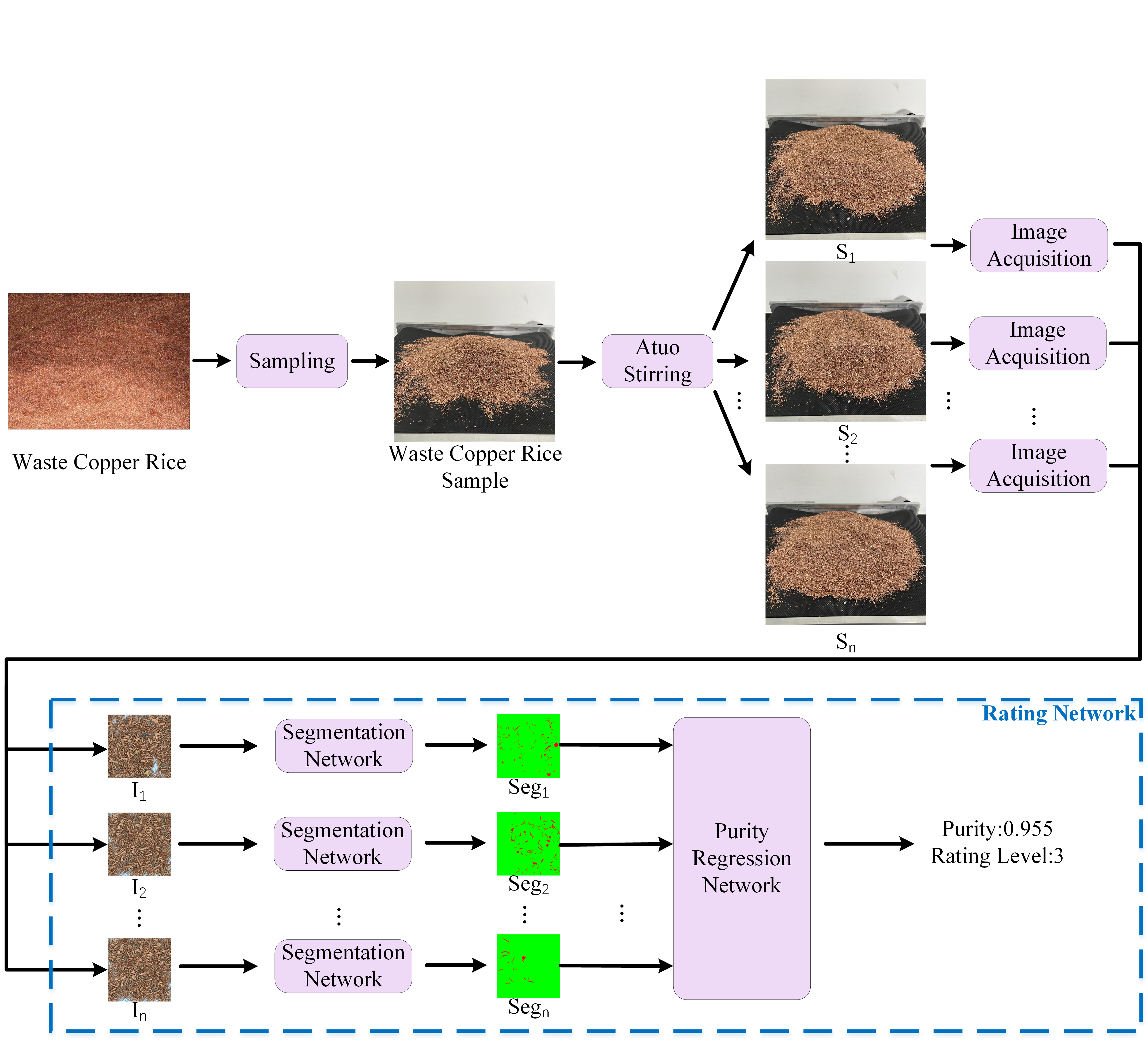}
{Our rating process.\label{fig3}}

In Fig. 3, we firstly sample the waste copper granules and stir samples n times like the existing manual process requires. Since our rating system works automatically and does not consume too much manpower, our sampling process can be repeated many times. Secondly, we capture images of stirred sample and convert the rating task into a 2D image recognition task. Finally, our rating network takes the captured n images as the input and outputs the purity and rating level of current sample. Our rating network consists two subnetworks, of which the segmentation subnetwork outputs the semantic segmentation heatmaps of n images successively, and the purity regression network calculates the purity and rating level according to the group of heatmaps. 

\subsection{Our Rating Network}
To rate the waste copper granules accurately and efficiently, we design a rating network based on convolutional neural network, and the overall structure of the network is shown as Fig. 4.
\Figure[t!](topskip=0pt, botskip=0pt, midskip=0pt)[width=1\textwidth]{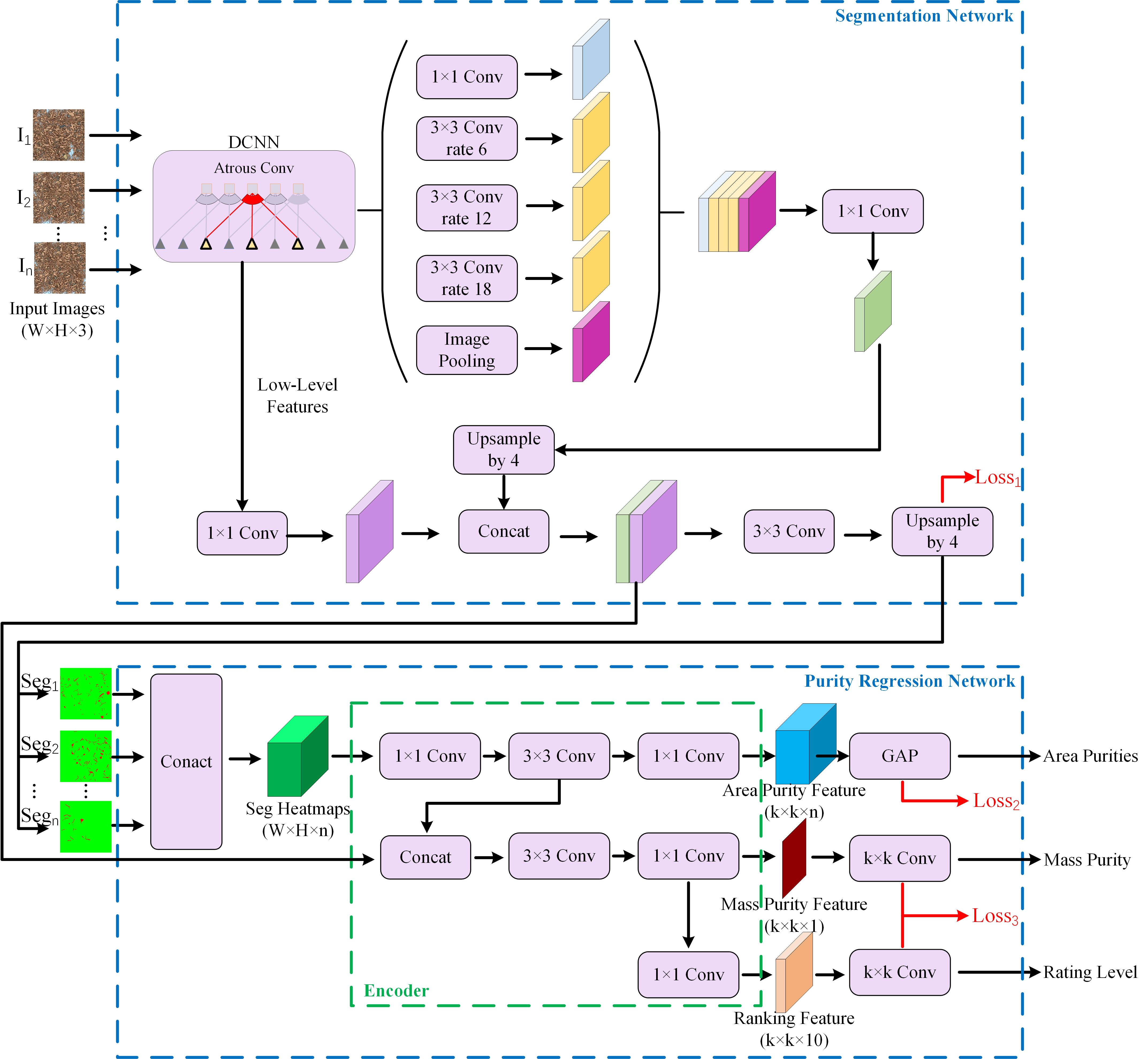}
{The structure of our rating network.\label{fig4}}

In Fig. 4, our rating network consists two subnetworks: segmentation network and purity regression network. The segmentation network is a well trained fully convolutional semantic segmentation model DeepLabV3+, which extracts deep features through down sampling encoder, and extracts shallow features through up sampling and outputs heatmap using a decoder. N images are fed into the segmentation network in turn to obtain the segmented heatmaps which classifies each pixel of the corresponding input image. Then, the purity regression network splices n heatmaps together as input and outputs area purities, mass purity, and rating level respectively. In the purity regression network, we design a three branches encoder to get area purity feature, mass purity feature, and ranking feature respectively. Finally, we use a global average pooling to get area purities of n heatmaps and use a simple convolution kernel which has same size as features to decode mass purity and rating level. 

We train the rating network in three phases to facilitate easier convergence of our rating model and increase accuracy. Firstly, we train the segmentation network in the way of deeplabv3+ to identify copper and impurities and freeze parameters of the segmentation network. Then, we train the area purity branch independently. Finally, we train the mass purity branch and the rating level branch jointly. In the inference stage, we feed n images of each sample into the rating network to calculate the mass purity and rating level. 

In the existing manual purity analysis and rating method, samples' purity is directly given by manual experience. This black box rating method is lack of objectivity and persuasiveness. In our machine vision based rating system, the rating task is converted to a computer vision task and the rating result is computed by our rating network. Each subnetwork of our network addresses specific subproblems, which enhances interpretability.

The segmentation subnetwork distinguishes between copper and non-copper in waste copper granules images and Fig. 5 shows some masked heatmaps produced by segmentation network. All impurity regions marked as blue in Fig. 5 are automatically recognized and other regions are filled with copper. The output heatmap has the same size as input image and each pixel is tagged with zero or one, denoting the presence of copper or an impurity, respectively.

\begin{figure}[htbp]
	\centering
	\includegraphics[width=0.5\textwidth]{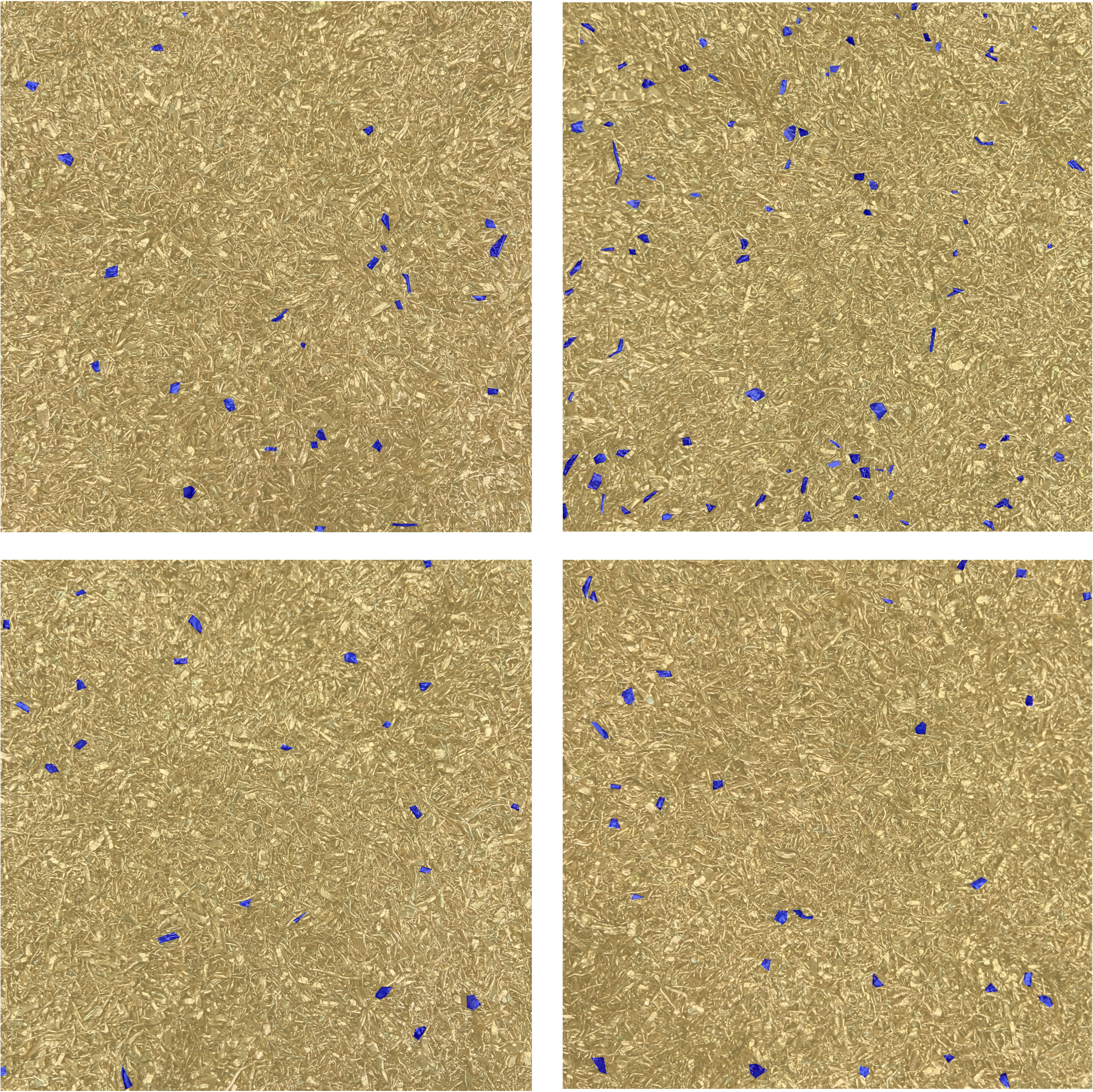}
	\caption{Masked heatmaps produced by segmentation network.}
\end{figure}

The purity regression subnetwork computes the area purity, mass purity and ranking level using heatmaps in an effective way.  Under the supervision of the area purity branch, the mass purity branch fuses the semantic features of the segmentation subnetwork and the area purity features and outputs the mass purity. The ranking branch classifies the rating level result based on other branches.

\section{Experiments}
In this section, we evaluate the proposed rating system on real waste copper granules data. Firstly, we build a  new dataset based on real waste copper granules. Then, we perform extensive experiments on the dataset and compare with ground truth and the best manual method to verify the effectiveness of the proposed system.

\subsection{Dataset}
Firstly, we simply sample the recycled waste copper granules for 220 times and split first 200 samples as training-validation datasets and the other 20 samples as the test dataset. Then, we purify each sample to get the real mass purity and rating level as ground truth of our experiments.

For each waste copper granules sample, we stir it and collect image data 128 times, which means n is 128 in Fig. 4. In these 128 times of acquisition processes, most of copper granules and impurities have the opportunity to appear on the top and be captured by the camera. Finally, we got 20480 training images, 5120 validation images and 2560 test images. For convenience, we divide every 20 samples  into 1 group and all the data are divided into 8 groups of training dataset, 2 group of validation dataset and 1 group of test dataset.

When generating ground truth for each sample, we label a mass purity scalar, a one-hot rating level vector, 128 semantic segmentation heatmaps and a 128 dimensional area purity vector. One sample with its annotations are shown as Fig. 6.

In Fig. 6, there are the sampled waste copper granules sample, 2D images after stirred, the corresponding n segmented binary heatmaps, the corresponding n area purity scalars, a mass purity scalar and a one-hot ranking level scalar from left to right. Heatmaps are manually marked with labelme, area purity is calculated according to the corresponding heatmap, the mass purity scalar and ranking level scalar are obtained by chemical purification.

\begin{figure}[htbp]
	\centering
	\includegraphics[width=0.5\textwidth]{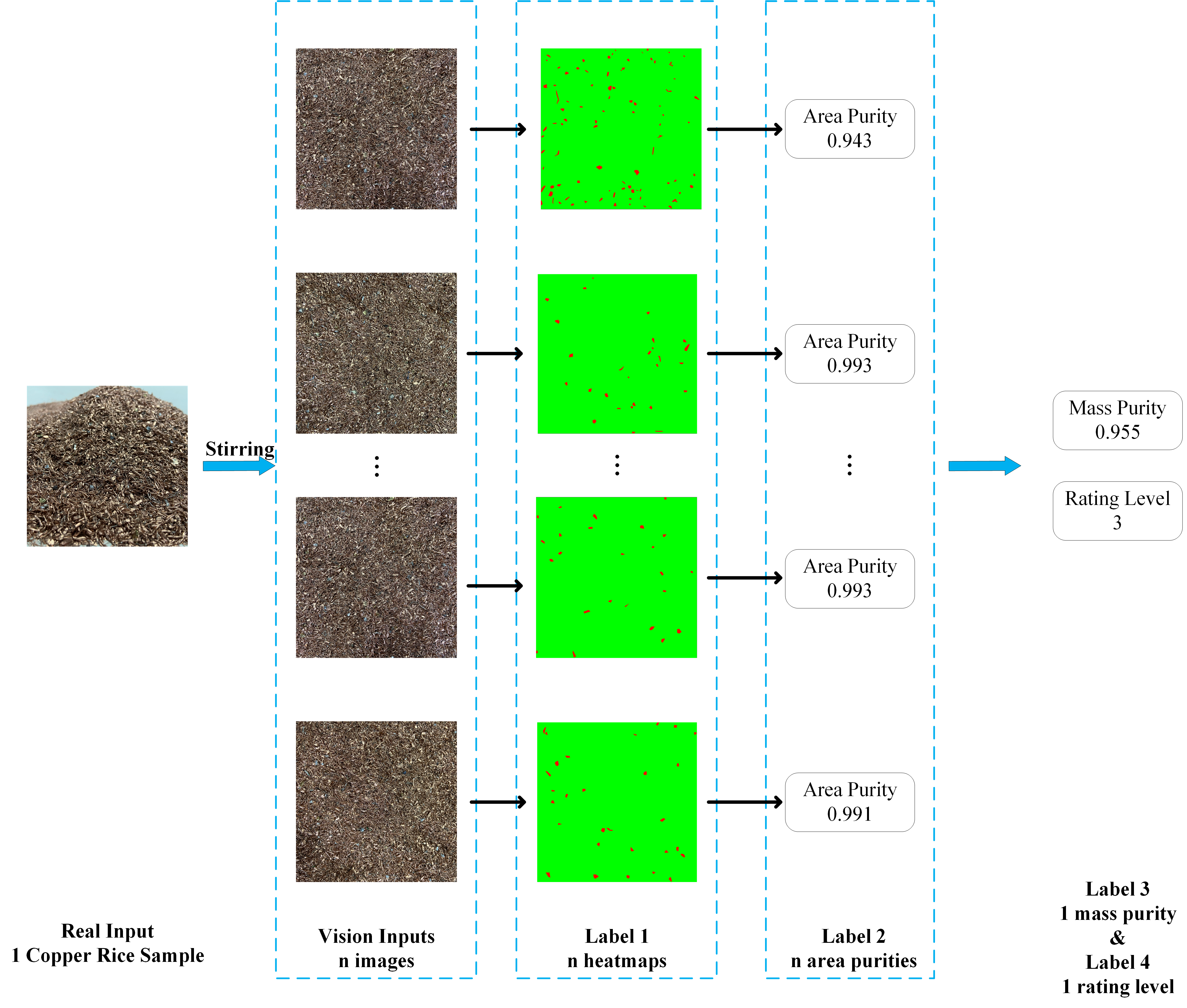}
	\caption{Real input copper granules sample and corresponding annotations.}
\end{figure}

\subsection{training}
We train our rating network in three steps since it is challenging to directly estimate the mass purity of waste copper granules.

The first step is to train the segmentation network. We convert the purity estimation task into a visual task, and it is necessary to distinguish between copper and impurities in the input image accurately. We train the segmentation network following the way of deeplabv3+ with Loss1:
\begin{equation}
Los{s_1}\left( {\hat y,y} \right) =  - \left( {1/K} \right)\sum\nolimits_{j = 1}^K {{y_j}log} {\hat y_j}
\label{eq2}\end{equation}
Loss1 in \eqref{eq2} is the average cross entropy loss of all pixels, and K is the number of pixels.

In order to improve the robustness and accuracy of our segmentation network, we expand training dataset by means of data augmentation. We propose a new augmentation approach based on the trait of our copper granules data in addition to the conventional semantic segmentation data augmentation methods like flip, rotation, and translation. Inspired by cutmix\cite{b34} and mosaic\cite{b35} method  used in classification and detection field, the proposed method comprises of three steps. In the first step, we extract and save all impurity regions from training set. In the second step, for each training image, we randomly select and rotate k extracted impurity regions. Finally, we paste k impurity regions into the image’s copper granules region and generate one new training image. The final mIoU\cite{b36} on training datasets is 0.904. After inferencing on validation images, the final mIoU is 0.878, indicating that our segmentation network has good generalization ability.

The mass purity cannot be estimated directly using a single pipeline whereas only the area purity can be calculated. Additionally, area purity has a lower fitting difficulty than mass purity. Therefore, we believe that the learning process of area purity is conducive to the learning of mass purity. So prior to considering mass purity branch, we train the area purity branch of purity regression network in the first place. After the segmentation network is well-trained, we freeze its parameters and use area purity of all training images to train the area purity branch according to Loss2 in Fig. 4. Loss2 is a simple L1 loss function as \eqref{eq3}.
\begin{equation}
Los{s_2}\left( {\widehat y,y} \right) = \left( {1/n} \right)\sum\nolimits_{i = 0}^n {\left| {{{\widehat y}^i} - {y^i}} \right|} 
\label{eq3}\end{equation}

The third step is to train the mass purity branch and ranking branch of the purity regression network. After training the segmentation subnetwork and area purity branch, we freeze the segmentation network and then train the whole purity regression network using Loss3 in Fig. 4. Loss3 is weighted by a L1 loss function and a focal loss function\cite{b37}, as shown in \eqref{eq4}, where we set $\alpha=0.5$.
\begin{equation}
Los{s_3}\left( {\widehat y,y} \right) = \alpha \left| {\widehat y - y} \right| + (1 - \alpha )FL\left( {\widehat y,y} \right)
\label{eq4}\end{equation}

After three-step training, our rating network can directly calculate the mass purity and rating level according to the input waste copper granules sample.

\subsection{Experimental results}
We evaluate the proposed rating network on real datasets and verify the effectiveness of our network by comparing and analyzing area purity, mass purity and rating level with ground truth and results given by experienced engineers.

\begin{figure}[htbp]
	\centering
	\includegraphics[width=0.5\textwidth]{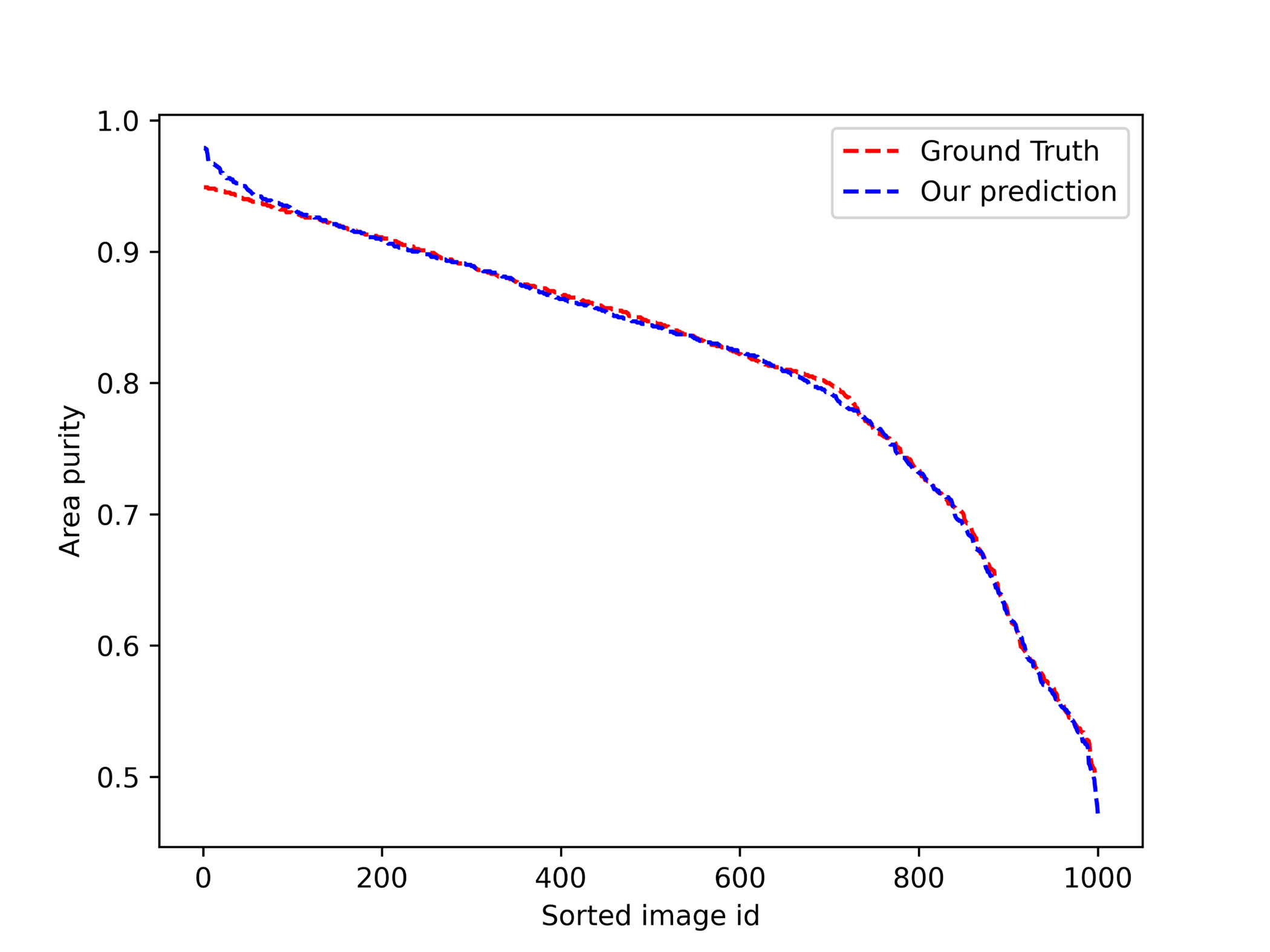}
	\caption{Area purity results of 1000 images.}
\end{figure}
Firstly, we calculate area purity on 1000 valuation images using our area purity branch and compare with ground truth. The comparison results are shown in Fig. 7.

In Fig. 7, we draw the area purity of 1000 images into a curve, where the blue curve represents our rating result, the red curve represents the real area purity. For convenience, we sort images in ascending order according to real area purity. Fig. 7 shows that the area purity branch of our rating network is reliable.

Then, we report the final mass purity and ranking results of the proposed rating system on the 10 groups of training-validation datasets in Table. 1 and Table. 2. The validation set's predicted copper granules mass purity is fairly close to ground truth, demonstrating the utility of our rating module on copper granules datasets. The ranking result on training-validation datasets match the ground truth exactly, which shows that our rating system is reliable.
\begin{table}[!ht]
\caption{Mass purity in training and validation datasets}
    \centering
    \begin{tabular}{|c|c|c|}
    \hline
        Mass Purity & Prediction & Ground Truth \\ \hline
        Training Dataset1  & 0.911 & 0.938 \\ 
        Training Dataset2 & 0.997 & 0.970 \\ 
        Training Dataset3 & 0.892 & 0.855 \\ 
        Training Dataset4 & 0.765 & 0.797 \\ 
        Training Dataset5 & 0.966 & 0.944 \\ 
        Training Dataset 6 & 0.940 & 0.908 \\ 
        Training Dataset 7 & 0.744 & 0.771 \\ 
        Training Dataset 8 & 0.811 & 0.842 \\ 
        Validation Dataset 1 & 0.792 & 0.751 \\ 
        Validation Dataset 2 & 0.691 & 0.735 \\ \hline
    \end{tabular}
    \label{tab1-1}
\end{table}

\begin{table}[!ht]
\caption{Rating level in training and validation datasets}
    \centering
    \begin{tabular}{|c|c|c|}
    \hline
        Rating Level & Prediction & Ground Truth \\ \hline
        Training Dataset1  & 3 & 3 \\
        Training Dataset2 & 1 & 1 \\
        Training Dataset3 & 4 & 4 \\
        Training Dataset4 & 6 & 6 \\
        Training Dataset5 & 2 & 2 \\
        Training Dataset 6 & 3 & 3 \\
        Training Dataset 7 & 6 & 6 \\
        Training Dataset 8 & 5 & 5 \\
        Validation Dataset 1 & 6 & 6 \\
        Validation Dataset 2 & 7 & 7 \\ \hline
    \end{tabular}
    \label{tab1-2}
\end{table}

The rating level on test dataset is shown in Table. 3. Although the ground truth of area purities and mass purity are unknow, the prediction rating level can still match. It shows that the proposed recognition model and rating system has generalization ability and our system is competent for rating task.

\begin{table}[!ht]
\caption{Rating result in test dataset}
    \centering
    \begin{tabular}{|c|c|c|c|}
    \hline
        ~ & Our mass purity & Our rating level & Real rating level \\ \hline
        Test dataset & 0.976 & 1 & 1 \\ \hline
    \end{tabular}
    \label{tab2}
\end{table}

Finally, in the actual rating scene, the number of ranking levels is all the matters, and the more ranking levels there are, the greater the challenge is for rating system. Therefore, we analyze the ranking error under different number of ranking levels and the error rate of 220 samples is shown as Fig. 8. In addition to our ranking result, engineers are invited with 10 years of rating experience and 1 year learning experience to rate each waste copper granules sample respectively. We call the experienced engineer ``teacher'' and the other engineer “student”.

\begin{figure}[htbp]
	\centering
	\includegraphics[width=0.5\textwidth]{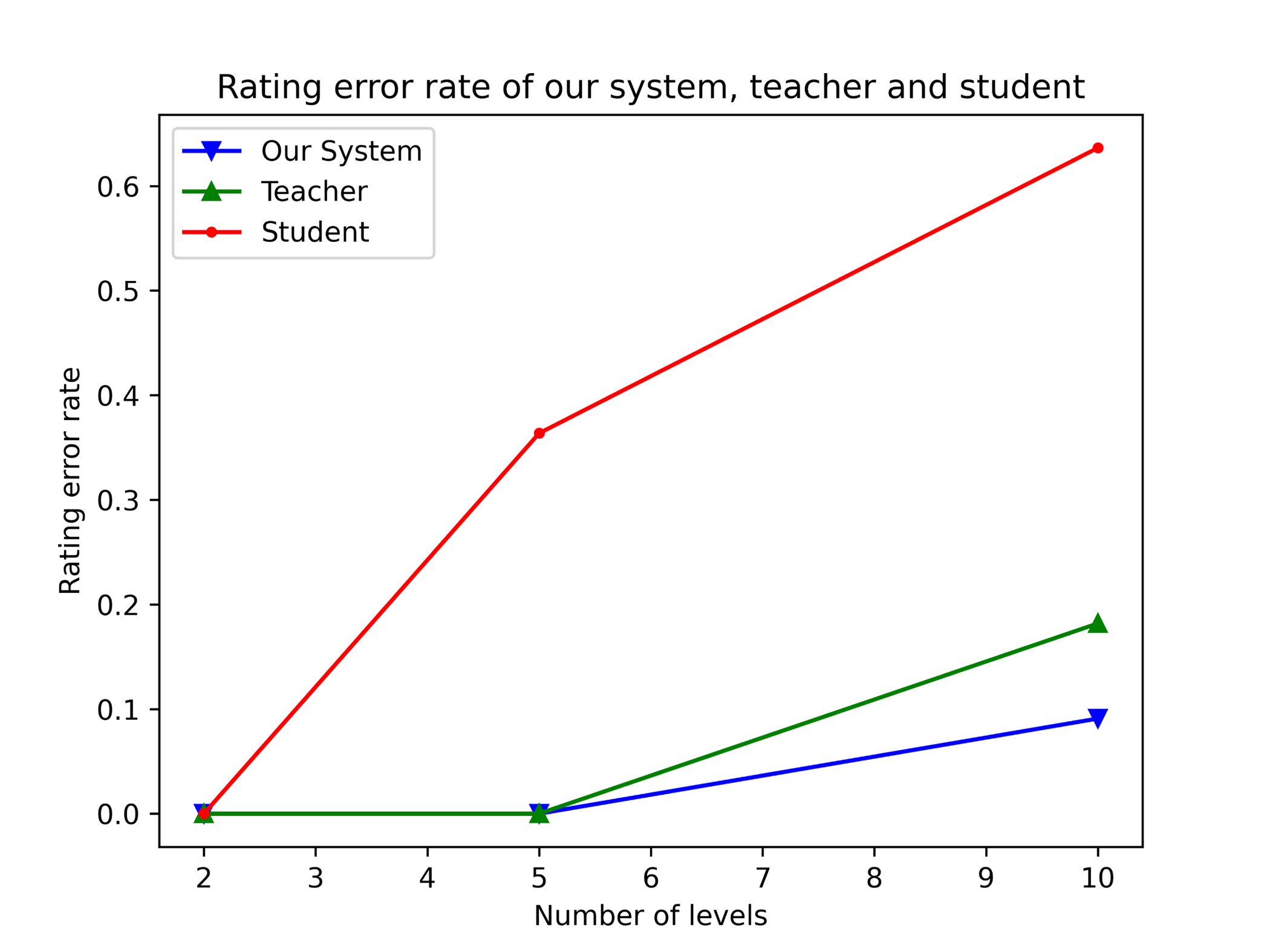}
	\caption{Rating error of our system, teacher and student.}
\end{figure}

In Fig. 8, Rating error rate of our system,teacher and students are indicated by blue, green, and red dots, respectively. If we divide purity into 2 levels, student can also correctly rate all waste copper granules. However the rate of student rating inaccuracy immediately rises as the number of levels increases. When the number of levels is 5, our system and teacher both correctly rate all waste copper granules and our system is more effective and stable if the number of ranking levels is 10.

In our experiments, we firstly split 220 samples into 11 groups, and get ground truth of each sample after chemical purifying. Then, we evaluate each group dataset separately and compare the results with those of skilled engineers to prove the effectiveness of our system. In real industrial scenarios, our machine vison based rating system can work consistently, steadily and effectively.

\section{Conclusions}
In the field of waste copper granules recycling, the current commonly used pricing method relies on experienced engineers to estimate the purity of waste copper granules and rate it. In this paper, we rethink copper granules purity from the perspective of semantic segmentation and propose a waste copper granules rating network based on machine vision, including a segmentation subnetwork and a purity regression subnetwork. To improve the rating accuracy, we develop a data augmentation method to enhance the generalization ability and precision of copper granules segmentation model. Furthermore, to ensure the effectiveness and accuracy of network, we train our rating network in three steps. Experiments on real waste copper granules dataset verify the effectiveness and superiority of the proposed network. Particularly, if the ranking level is split into more hierarchies, our approach will highly outperform the manual technique in terms of effectiveness, reliability, and objectivity.

\EOD

\end{document}